\documentclass[10pt,twocolumn,letterpaper]{article}

\usepackage{cvpr}              

\definecolor{cvprblue}{rgb}{0.21,0.49,0.74}
\definecolor{yellowhl}{RGB}{255,248,184} 
\definecolor{redhl}{RGB}{255,217,217}    
\definecolor{orangehl}{RGB}{255,235,204} 
\usepackage[pagebackref,breaklinks,colorlinks,allcolors=cvprblue]{hyperref}

\usepackage[utf8]{inputenc}

\usepackage{multirow}
\usepackage{tabularx}
\usepackage{colortbl} 
\usepackage{xcolor}
\usepackage{rotating} 
\usepackage{hhline} 
\usepackage{float}
\usepackage{afterpage}
\usepackage{subcaption}
\usepackage{booktabs} 
\usepackage{amssymb}  
\usepackage{url} 
\usepackage{pifont}
\usepackage{marvosym} 

\def\paperID{5470} 
\def\confName{CVPR}
\def\confYear{2026}

\title{FilterGS: Traversal-Free Parallel Filtering and Adaptive Shrinking \\for Large-Scale LoD 3D Gaussian Splatting}

\author{
  Yixian Wang \quad
  Haolin Yu \quad
  Jiadong Tang \quad
  Yu Gao \quad
  Xihan Wang \quad
  Yufeng Yue \quad
  Yi Yang$^{\dagger}$
  \\
  Beijing Institute of Technology \quad
  $^\dagger$ Corresponding Author
}

\newcommand{\heading}[1]{\vspace{6pt}\noindent\textbf{#1}\hspace{1em}\ignorespaces}

\begin{document}

\twocolumn[{%
\renewcommand\twocolumn[1][]{#1}%
\maketitle
\begin{center}
   \hsize=\textwidth 
   \centering
   \includegraphics[width=\linewidth]{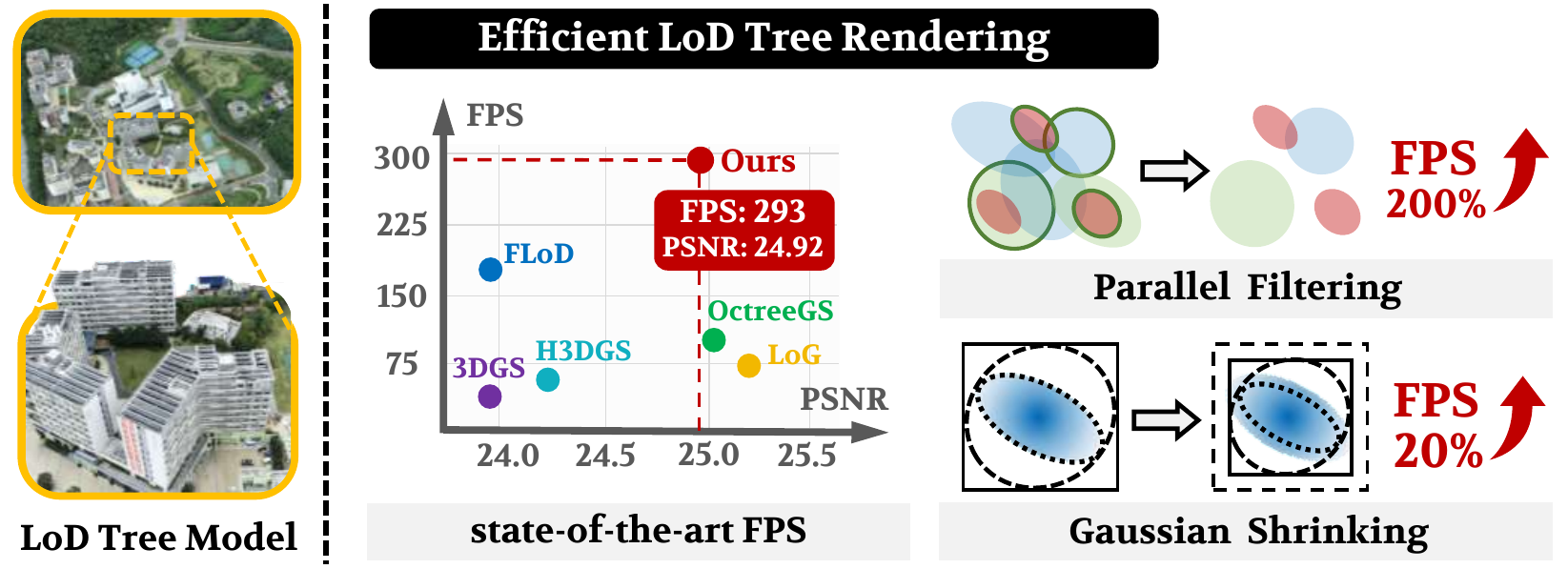}
   \captionof{figure}{\textbf{FilterGS} enables efficient rendering of large-scale 3D Gaussian Splatting LoD trees. By integrating parallel filtering and Gaussian shrinking strategies, our method achieves an average rendering speed of nearly 300 FPS across 6 scenes. It significantly outperforms representative state-of-the-art approaches while maintaining competitive reconstruction quality.}
   \label{img:3-1}
\end{center}%
}]

\begin{abstract}

3D Gaussian Splatting has revolutionized neural rendering with real-time performance. However, scaling this approach to large scenes using Level-of-Detail methods faces critical challenges: inefficient serial traversal consuming over 60\% of rendering time, and redundant Gaussian-tile pairs that incur unnecessary processing overhead. To address these limitations, we introduce FilterGS, featuring a parallel filtering mechanism with two complementary filters that select Gaussian elements efficiently without tree traversal. Additionally, we propose a novel GTC metric that quantifies the redundancy of Gaussian-tile key-value pairs. Based on this metric, we introduce a scene-adaptive Gaussian shrinking strategy that effectively reduces redundant pairs. Extensive experiments demonstrate that FilterGS achieves state-of-the-art rendering speeds while maintaining competitive visual quality across multiple large-scale datasets. Project page:
\url{https://github.com/xenon-w/FilterGS}

\end{abstract}
    
\section{Introduction}

3D reconstruction and neural rendering powered by 3D Gaussian Splatting (3DGS)\cite{3dgs} have demonstrated significant utility in applications such as autonomous driving\cite{splatam, opengsslam}, augmented reality\cite{instant3d,dronesplat}, and the metaverse\cite{opengsfusion, Gaussiangraph}. However, despite being a fundamental requirement for these applications, real-time rendering of large-scale scenes continues to face significant challenges.

Practical large-scale scenes typically demand an extremely large number of Gaussian primitives to faithfully represent fine-grained geometry and appearance. Directly rasterizing all these Gaussians without any filtering or simplification incurs prohibitive computational overhead and excessive memory consumption, which drastically slows down the rendering speed and makes it difficult to meet real-time constraints.
Although 3DGS significantly accelerates rendering through its differentiable rasterization pipeline, rendering large-scale  scenes requires managing a growing number of Gaussians and tiles\cite{flashgs}, leading to significantly slower rendering speeds in large-scale scenes and limiting its practical adoption.


To accelerate 3DGS rendering in large-scale scenes, a series of recent methods~\cite{octreegs, hieGS, flod, LoG} have introduced Level-of-Detail (LoD) strategies. As a widely adopted technique for efficient rendering, LoD aims to dynamically adjust the complexity of scene representation according to the current viewpoint. The core idea is to select and render only a subset of primitives that satisfy certain view-dependent criteria, so as to restrict the number of Gaussians processed in each frame and keep the rendering computation within a manageable range. In this way, LoD schemes effectively alleviate the excessive computational burden caused by massive Gaussians in large-scale scenarios, making real-time rendering more feasible. Nevertheless, we observe that existing LoD-based approaches still suffer from several critical inefficiencies that hinder their practical performance.

The primary challenge is \textit{inefficient hierarchy traversal and Gaussian processing}. To adequately capture scene details, LoD-based methods employ hierarchical Gaussian representations. However, the associated processes introduce a significant performance bottleneck. Some methods\cite{hieGS,flod,lodge} rely on hierarchical traversal to select suitable Gaussians for the current frame from the Gaussian tree, where layer-wise searching incurs substantial time costs. Other methods\cite{octreegs, scaffoldgs} adopt a neural-Gaussian design that requires retrieving attributes of neighboring Gaussians from selected key nodes, which further slows down the entire rendering pipeline. 
In fact, these operations alone can consume more than 60\% of the total rendering time in some test scenes. 

The second challenge is \textit{redundant Gaussian-tile pairs}. The drive for detailed representation necessitates multi-level hierarchical Gaussian structures, yet the preprocess of rendering forms a vast number of redundant Gaussian-tile pairs. Despite being skipped during $\alpha$-blending, these redundant pairs impose the full computational overhead of sorting and processing, thereby degrading rasterization speed. Existing efforts to mitigate such redundancy include fixed-threshold Gaussian shrinking strategies\cite{flashgs}, which fail to sufficiently reduce redundant pairs in large-scale scene settings. Meanwhile, learning-based methods\cite{3DGSsparser, NoRedundancy} show promise for pruning redundant pairs in standard 3DGS, but their per-Gaussian trained shrinking coefficients introduce level-switching artifacts and overfitting when adapted to hierarchical LoD structures.

To address these issues, we propose FilterGS, featuring two key innovations: a parallel filtering mechanism that enables Gaussian selection without level-by-level traversal through two complementary filters. Additionally, we introduce a scene-adaptive Gaussian shrinking strategy that proactively reduces redundant Gaussian-tile pairs via GTC-based rational scaling. Together, our parallel filters and shrinking strategy form a cohesive framework that achieves efficient rendering in large-scale scenes. Our contributions can be summarized as follows:

\begin{itemize}

\item We propose FilterGS, a comprehensive rendering acceleration method that achieves state-of-the-art rendering speed through efficient filtering and redundant key-value pair elimination, while maintaining excellent reconstruction quality.

\item We introduce two parallel filters tailored for the LoD Gaussian model that enable rapid hierarchical filtering, boosting rendering FPS by over 200\% and effectively resolving critical bottlenecks in the rendering pipeline.

\item We introduce a universal scene-adaptive Gaussian shrinking strategy that dynamically adjusts pruning aggressiveness to eliminate redundant Gaussian-tile pairs, delivering an improvement of over 20\% in FPS while preserving perceptually critical structures.

\end{itemize}

\section{Related Work}

\heading{Level-of-Detail (LoD).} 3D Gaussian Splatting (3DGS)\cite{3dgs} achieves high-fidelity novel view synthesis and real-time rendering, yet scales poorly when applied to large environments due to the exponential growth of Gaussian primitives.  

\begin{figure*}[t]
   \centering
   \includegraphics[width=\linewidth]{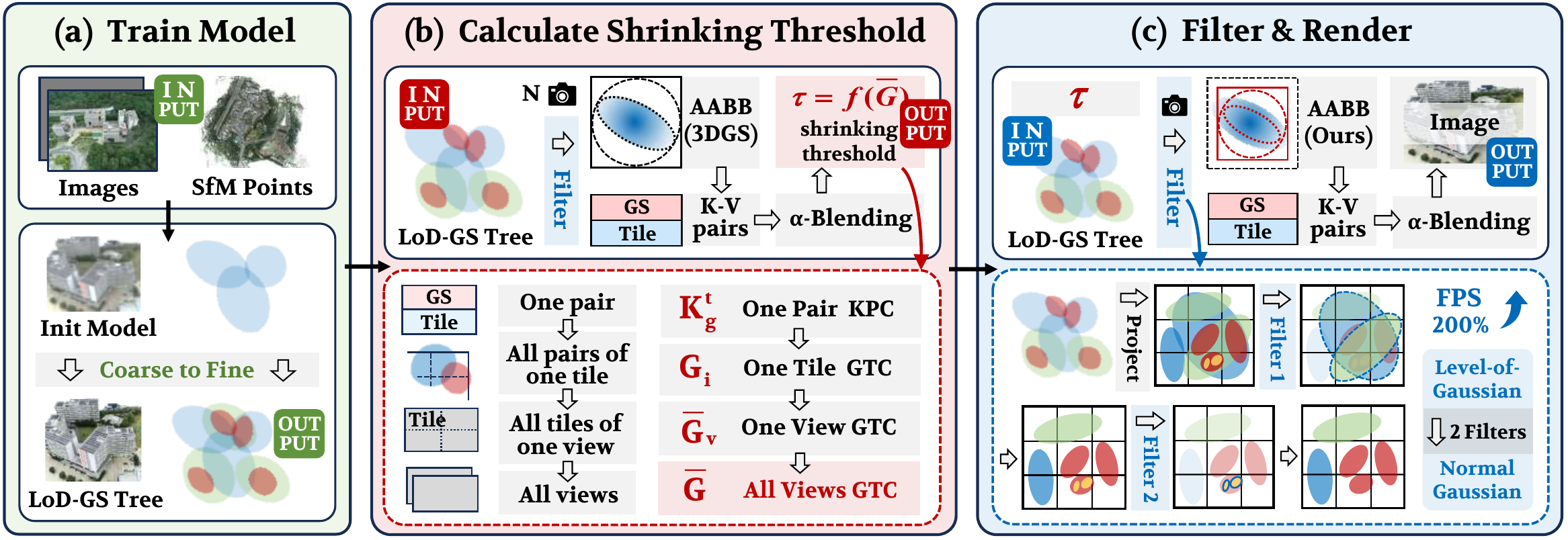}
   \caption{\textbf{The framework of FilterGS}. (a) Given a set of images and SfM points for a large-scale scene, we first train a LoD-GS tree model. (Sec.~\ref{chap:3.1}) (b) To quantify redundancy of scene-wide Gaussian-tile pairs, we perform a pre-rendering pass over all training views. While following the standard 3DGS pipeline, this stage computes the GTC metric $\bar{G}$ and derives a shrinking threshold $\tau=f(\bar{G})$. (Sec.~\ref{chap:3.3}) (c) In the formal rendering stage, the pre-computed scaling factor $\tau$ and the tree model are processed by two specialized filters to efficiently select Gaussians for rendering. These Gaussians are adaptively scaled by $\tau$ during AABB formation, significantly reducing redundant key-value pairs before final sorting and $\alpha$-blending. (Sec.~\ref{chap:3.4})}
   \label{img:3-1}
\end{figure*}

To handle large-scale scenes, several works \cite{vastGaussian,cityGaussian,cityGaussianv2,doGaussian,automated} adopt chunk-based training strategies inspired by large-scale NeRF systems \cite{UrbanNerf,GridNerf,switchNerf,blockNerf,megaNerf,mcnerf}, which effectively reduce GPU memory usage but result in a large number of Gaussians, thus limiting the rendering FPS. 

A more flexible paradigm is the Level-of-Detail (LoD) strategy, which dynamically adjusts Gaussian density and resolution according to scene complexity or perceptual relevance. Hierarchical-GS\cite{hieGS} first introduces a continuous LoD mechanism for large-scale Gaussian rendering, balancing detail and efficiency. Octree-GS\cite{octreegs} and FLoD\cite{flod} extend this concept using discrete hierarchical structures, progressively refining coarse Gaussians into finer ones to recover detailed geometry. LODGE\cite{lodge} further implements dynamic block loading, where only active Gaussians are streamed to memory, enabling real-time rendering even on mobile platforms.  However, existing LoD-based systems still suffer from heavy Gaussian–tile construction and filtering overheads, causing pre-rasterization redundancy and limiting rendering efficiency.

\heading{Gaussian Rendering Acceleration.} When processing large-scale Gaussian elements or applications demanding real-time performance, the vanilla 3DGS pipeline often becomes a critical bottleneck, motivating extensive research on rendering acceleration. Early methods\cite{minisplat,Gaussianspa} employ lightweight representations and point-culling schemes to reduce computational overhead, while more recent approaches, such as Potamoi\cite{potamoi}, GS-Cache\cite{gscache}, FlashGS\cite{flashgs}, TC-GS\cite{tcgs} and SpeedySplat\cite{speedysplat}, further optimize rasterization through inter-frame reuse, preprocessing, and careful pipeline reorganization. Nonetheless, the rasterizer continues to process numerous negligible Gaussians, leading to substantial computation waste, thereby highlighting the need for a filtering mechanism that selectively retains only the most relevant pairs for efficient rendering.

To reduce redundant key-value pairs, FlashGS\cite{flashgs} leverages a fixed threshold to shrink Gaussians. However, this approach has proven inadequate in large-scale scenes, as it still generates a considerable number of redundant pairs. While learning-based approaches \cite{3DGSsparser, NoRedundancy} have demonstrated remarkable efficacy in eliminating redundant key-value pairs for vanilla 3DGS, training shrinking coefficients on a per-Gaussian basis introduces LoD-switching artifacts and overfitting issues when applied to hierarchical representations. In contrast to vanilla 3DGS, LoD-3DGS is characterized by tree architectures with drastically varying node sizes across different levels, exacerbating such limitations.
\section{Method}
\subsection{Preliminary}
\label{chap:3.1}

\heading{3D Gaussian Splatting (3DGS).} 3DGS\cite{3dgs} represents scenes explicitly through anisotropic 3D Gaussian primitives and enables efficient rendering through differentiable splatting. A Gaussian primitive is defined by the following set of parameters: a center position $\mu$, a covariance matrix $\Sigma$, an opacity value $\alpha$, and spherical harmonics coefficients representing color $c$. The influence of each Gaussian on a 3D point $x$ can be described: 
\begin{equation}
    G(x) = e^{-\frac{1}{2}(x-\mu)^\top \Sigma^{-1}(x-\mu)}
    \label{3dgs}
\end{equation}

To render image, 3DGS first project Gaussian primitives onto the 2D plane. Specifically, the color $C$ of a pixel is computed by blending $N$ ordered 2D Gaussians overlapping that pixel:
\begin{equation}
    C = \sum_{i \in N} c_i \alpha_i T_i, \qquad T_i=\prod_{j=1}^{i-1} (1 - \alpha_j)
    \label{3dgs-color}
\end{equation}
where $c_i$ and $\alpha_i$ represent the color and opacity of the projected 2D Gaussian elements. $T_i$ represent the residual transmittance.

\begin{figure*}[t]
   \centering
   \includegraphics[width=\linewidth]{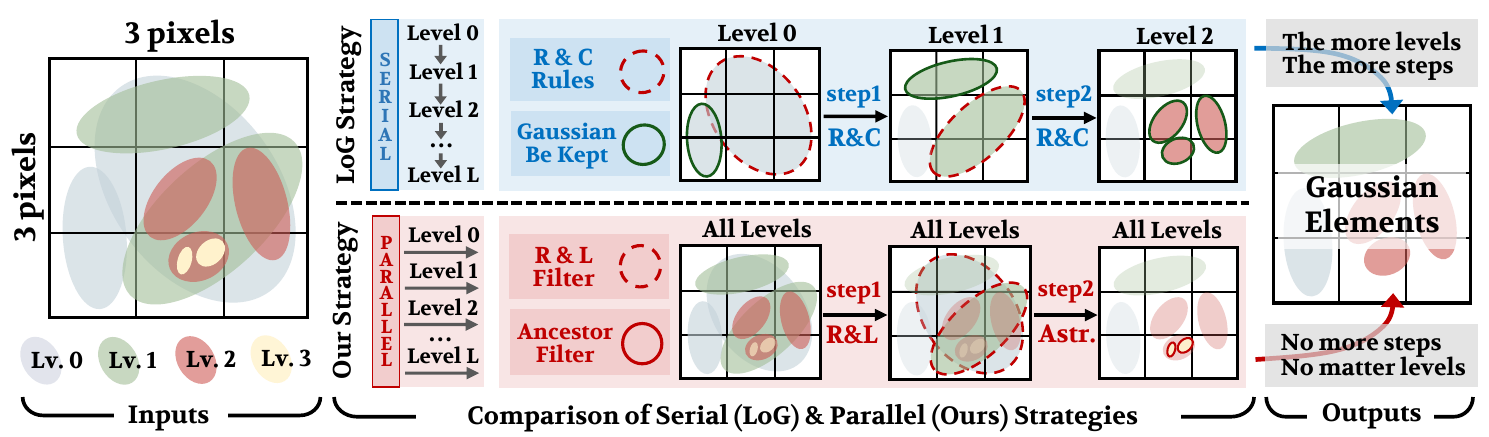}
   \caption{\textbf{Comparison of serial and parallel filtering strategies.} LoG starts from level=0 and progressively applies R\&C rules to filter internal nodes with oversized 2D radii and child nodes, retaining only leaves or sufficiently small internal nodes. In contrast, FilterGS simultaneously applies the R\&L Filter across all levels to preserve nodes with small 2D radii and all leaf nodes, while introducing an Ancestor Filter to exclude nodes whose ancestor paths contain qualified internal nodes. This yields a key advantage: while LoG’s serial filtering requires steps proportional to tree depth (more levels, more steps), FilterGS completes parallel filtering in just two steps (No more levels, no matter levels).}
   \label{img:3-1}
\end{figure*}

\heading{Hierarchical Gaussian Tree.} Some LoD-based 3DGS pipelines\cite{LoG,flod} organize scenes as hierarchical Gaussian tree model, where each node encapsulates a full Gaussian primitive. The key advantage is LoD selection yields a set of standard Gaussians with complete parameters, enabling direct rasterization without additional pre-rendering decoding
The filtered Gaussians already satisfy the interface of vanilla 3DGS rasterizers, establishing an ideal foundation for aggressive rendering acceleration.

Our implementation follows the canonical recursive construction of such trees. For a parent node $v$ with rotation matrix $R(\mathbf{q}_v)$ and anisotropic scale $\mathbf{s}_v$, we place $K_v$ children using oriented offsets $\mathbf{o}_k\in \{ \pm \frac{s_{x,v}}{2},\pm \frac{s_{y,v}}{2}, \pm \frac{s_{z,v}}{2} \}$: $\boldsymbol{\mu}_{u_k}=\boldsymbol{\mu}_v+R(\mathbf{q}_v)\mathbf{o}_k$. Each child shrinks geometrically, $\mathbf{s}_{u_k}=\gamma~\mathbf{s}_v$ with $0<\gamma<1$. To maintain appearance consistency while refining geometry, every child simply inherits the parent’s orientation, color, and opacity, i.e., $\mathbf{q}_{u_k}=\mathbf{q}_v$, $\mathbf{c}_{u_k}=\mathbf{c}_v$, and $\alpha_{u_k}=\alpha_v$. This yields a natural coarse-to-fine hierarchy whose nodes remain immediately renderable.

\subsection{Traversal-Free Parallel Filtering}
\label{chap:3.4}

To rapidly extract renderable Gaussians from the hierarchical structure, we project all Gaussians inside the view frustum into 2D Gaussians and cascade two complementary filters—\textit{R\&L Filter} followed by \textit{Ancestor Filter}. The former retains every leaf node and the nodes whose footprint is sufficiently small; the latter traverses ancestor paths to cull redundant nodes so that each branch finally keeps only the most suitable level.

\heading{Radius and Leaf (R\&L) Filter.} 
For every in-frustum Gaussian we compute the 2D covariance $\Sigma_{2D}$; its eigenvalues $\sigma$ represent the standard deviations along the screen-space principal axes. We define the pixel radius as $R_{2D} = 3\sigma$ and compare it against a threshold $\tau_R$(we set it to 3). Any node with $R_{2D} \le \tau_R$ is precise enough for the current frame and is retained. Different values of $\tau_R$ correspond to different sizes of the selected Gaussian primitives. Within a reasonable range, a smaller $\tau_R$ indicates finer Gaussian primitives, which in turn leads to clearer details in the rendered images.

However, directly applying this pruning criterion may incorrectly remove entire branches composed of Gaussians with uniformly large radii. This would result in undesirable holes in the rendered image. To address this issue, we exempt leaf nodes from the pruning threshold. Specifically, every leaf node is preserved at least once, ensuring that each branch maintains at least one representative node.

\heading{Ancestor Filter.} 
We first formalize the LoD Gaussian representation: let $N_{i,j}$ denote the $j$-th node at level $i$, where the level index $i$ ranges from 0 to $L$ (the maximum depth of the LoD tree). Each node $N_{i,j}$ is associated with an ancestor path $AP_{i,j}$, which is a ordered list containing the serial numbers of all ancestor nodes of $N_{i,j}$ (from the direct parent node up to the root node of the tree), enabling traceability of the complete ancestor lineage for any node in the hierarchy.

Because R\&L filtering may retain multiple nodes $N_{i,j}$ across different levels $i$ on the same branch (e.g., both an internal node $N_{i_1,j_1}$ and its descendant node $N_{i_2,j_2}$ with $i_2>i_1$ satisfy the radius threshold $\tau_R$), we precompute the ancestor path $AP_{i,j}$ for every node $N_{i,j}$ prior to rendering, where $AP_{i,j}$ explicitly enumerates the serial numbers of all ancestor nodes from the direct parent up to the root node. The ancestor filter leverages $AP_{i,j}$ for top-down redundancy removal: once an internal node $N_{i*,j*}$ passes R\&L filtering, all its descendant nodes $N_{i,j}$ ($N_{i*,j*} \in AP_{i,j}$) — including those provisionally kept as leaves — are traversed via their respective $AP_{i,j}$ and culled.

Ancestor Filter ensures that Gaussian nodes are retained only in branches where no ancestor node $N_{i',j'}$ ($i'<i_{\text{leaf}}$) meets the radius criterion $\tau_R$, thereby avoiding the rasterization of unnecessary multi-level Gaussian nodes on the same branch. This mechanism guarantees that branches with a valid coarser node ultimately retain just a single, highest-level Gaussian node.

\heading{Parallel Filtering Mechanism.} The hierarchical filtering in prior works\cite{LoG,flod} imposes a severe algorithmic constraint due to its strict level-wise dependency. The serial scheme requires $ T_{\text{serial}} $:
\begin{equation}
  T_{\text{serial}} = \sum_{\ell=0}^{L-1} \Big( T_{\text{calcu.}}(n_{\ell}) + T_{\text{synch.}} \Big)
\end{equation}
where $T_{\text{synch.}}$ denotes the fixed overhead of kernel launch and synchronization per level, and $T_{\text{calcu.}}(n_{\ell})$ corresponds to the time for radius estimation and filtering of all $n_{\ell}$ nodes at level $\ell$. The bottleneck of this serial scheme lies in the $\mathcal{O}(L)$ dependency on hierarchical depth $L$, precluding parallel processing and introducing significant interlayer synchronization overhead.

FilterGS fundamentally redefines the culling paradigm by abandoning layer-wise traversal. We implement the R\&L and Ancestor Filters within a unified CUDA kernel optimized for massive parallel processing. This architectural shift liberates the filtering time from the depth dependency. For all $N$ in-frustum Gaussians, the filtering time can be expressed as:
\begin{equation}
  T_{\text{parallel}} = T_{\text{synch.}} + T_{\text{calcu.}}(N)
\end{equation}

This traversal-free paradigm achieves two key algorithmic advantages: ~\textit{1) Depth Decoupling}: The computational cost of filtering scales only with the total number of Gaussians $N$, and is independent of the hierarchical depth $L$. \textit{2) Maximized Concurrency}: By processing all levels collectively, FilterGS fully exploits GPU SIMD execution and memory coalescing, allowing the entire filtering procedure to add only a marginal, constant overhead, even for tens of millions of Gaussians.

\subsection{Scene-Adaptive Gaussian Shrinking}
\label{chap:3.3}
3DGS utilizes Gaussian-tile key-value pairs to manage spatial relationships. However, the multi-level structure inherent in LoD-based methods introduces significant redundancy in the rasterization front-end. We find that over 70\% of these pairs contribute negligibly to the final rendering, making subsequent operations a pipeline bottleneck. To this end, we propose an adaptive Gaussian shrinking strategy to minimize these redundant pairs.

\heading{Key-value Pair Contribution (KPC).} 
To quantify the redundancy of Gaussian-tile key-value pairs, we introduce the \textit{Key-value Pair Contribution} (KPC). Our key insight is that the contribution of a Gaussian to the final image can be approximated by the expected number of pixels it influences, weighted by its blended opacity. For a key-value pair $g_k\rightarrow t_i$, its KPC is defined as:
\begin{equation}
    K_{g_k}^{t_i} = \sum_{j=1}^{B_x\times B_y}\alpha_{ij} T_{ij}
    \label{eqa:KPC}
\end{equation}
where $B_x\times B_y$ is the number of pixels in tile $t_i$, $\alpha_{ij}$ is the opacity of Gaussian $g_k$ at pixel $j$, and $T_{ij}$ is the transmittance before $g_k$ at that pixel. We define key-value pairs with contribution (KPC) below 0.01 as redundant.

$K_{g_k}^{t_i}$ can be interpreted as the effective number of pixels that $g_k$ contributes to within $t_i$, making it a natural and effective metric for quantifying the utility of a key-value pair.

\vspace{-1.0em}
\heading{Gaussian to Tile Contribution (GTC).}\label{chap:gtc}
To quantify the average contribution of Gaussians to one single tile, we define the \textit{Gaussian to Tile Contribution} (GTC). For a tile $t_i$, its GTC is computed as the average KPC of all Gaussians that influence the tile:
\begin{equation}
    G_{i} = \frac{1}{n_{\text{gs}}}~\sum_{j=1}^{n_{\text{gs}}}K_{g_j}^{t_i}
    \label{eqa:GTC}
\end{equation}
where $n_{\text{gs}}$ is the number of Gaussians that influence tile $t_i$, and $K_{g_j}^{t_i}$ denotes the KPC of the $j$-th Gaussian ($g_j$) for tile $t_i$ (as defined in Eq.~\ref{eqa:KPC}).

\begin{figure}[bp]
   \centering
   \includegraphics[width=\linewidth]{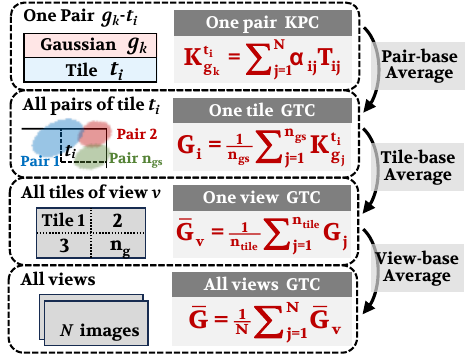}
   \caption{\textbf{Schematic Diagram of the Progressive Relationship Between the Four Core Formulas (Eqa.~\ref{eqa:KPC}, \ref{eqa:GTC}, \ref{eqa:GTC-view}, \ref{eqa:define-tau}) of Sec.~\ref{chap:3.3}.}}
   \label{img:3-3}
\end{figure}

$G_i$ reflects the average effective pixel contribution per Gaussian within tile $t_i$. A low $G_i$ value indicates Gaussian congestion—an inefficient rasterization state where a tile contains many Gaussians with negligible individual contributions. Such congestion typically arises from over-clustering and redundant primitives that add excessive computational overhead. In contrast, a high total KPC for a tile corresponds to genuinely translucent regions (e.g., leaves, fences) where low-opacity Gaussians are necessary for visual fidelity. This distinction is critical: GTC acts as a scene-aware metric that differentiates geometric redundancy from necessary low-opacity regions.

To obtain a view-level metric for overall Gaussian congestion across the entire image, we compute the average GTC over all $n_{\text{tile}}$ tiles in the view:

\begin{equation}
    \bar{G}_v = \frac{1}{n_{\text{tile}}} \sum_{j=1}^{n_{\text{tile}}} G_j
    \label{eqa:GTC-view}
\end{equation}

This averaging scheme ensures that each tile contributes equally to the assessment of the scene's overall efficiency, preventing the metric from being dominated by a small number of tiles with extremely high Gaussian density.

\begin{table*}[htbp]
    \centering
    \fontsize{9}{11}\selectfont
    \setlength{\tabcolsep}{3pt}
    \renewcommand{\arraystretch}{1.1}

\begin{tabularx}{\textwidth}{c|*{5}{>{\centering\arraybackslash}X}|*{5}{>{\centering\arraybackslash}X}|*{5}{>{\centering\arraybackslash}X}}
\noalign{\hrule height 0.8pt}

\multirow{2}{*}{\textbf{Method}} & 
\multicolumn{5}{c|}{\textbf{Block Small}~\cite{matrixcity}} & 
\multicolumn{5}{c|}{\textbf{Sci-Art}~\cite{urbanscene}} & 
\multicolumn{5}{c}{\textbf{Residence}~\cite{urbanscene}} \\

 & \rotatebox{0}{\fontsize{8}{10}\selectfont PSNR$\uparrow$} 
 & \rotatebox{0}{\fontsize{8}{10}\selectfont SSIM$\uparrow$} 
 & \rotatebox{0}{\fontsize{8}{10}\selectfont LPIPS$\downarrow$} 
 & \rotatebox{0}{\fontsize{9}{10}\selectfont $t_\text{f}$ $\downarrow$} 
 & \rotatebox{0}{\fontsize{8}{10}\selectfont FPS$\uparrow$} 
 & \rotatebox{0}{\fontsize{8}{10}\selectfont PSNR$\uparrow$} 
 & \rotatebox{0}{\fontsize{8}{10}\selectfont SSIM$\uparrow$} 
 & \rotatebox{0}{\fontsize{8}{10}\selectfont LPIPS$\downarrow$} 
 & \rotatebox{0}{\fontsize{9}{10}\selectfont $t_\text{f}$ $\downarrow$}
 & \rotatebox{0}{\fontsize{8}{10}\selectfont FPS$\uparrow$} 
 & \rotatebox{0}{\fontsize{8}{10}\selectfont PSNR$\uparrow$} 
 & \rotatebox{0}{\fontsize{8}{10}\selectfont SSIM$\uparrow$} 
 & \rotatebox{0}{\fontsize{8}{10}\selectfont LPIPS$\downarrow$} 
 & \rotatebox{0}{\fontsize{9}{10}\selectfont $t_\text{f}$ $\downarrow$} 
 & \rotatebox{0}{\fontsize{8}{10}\selectfont FPS$\uparrow$} \\
\hline

3DGS & 25.42 & 0.743 & 0.446 & -- & 58.0 & 22.35 & 0.717 & 0.311 & -- & 61.0 & 21.92 & 0.767 & 0.298 & -- & 62.0 \\
H3DGS & 25.73 & 0.756 & 0.440 & 5.51 & 83.0 & 22.41 & 0.719 & 0.312 & 53.46 & 72.0 & 21.99 & 0.769 & 0.294 & 7.12 & 38.0 \\
LoG & \cellcolor{redhl}{26.52} & \cellcolor{orangehl}{0.770} & \cellcolor{redhl}{0.414} & 10.46 & 77.0 & \cellcolor{orangehl}{23.27} & \cellcolor{yellowhl}{0.737} & \cellcolor{orangehl}{0.305} & 11.6 & 67.0 & \cellcolor{redhl}{22.7} & \cellcolor{redhl}{0.812} & \cellcolor{redhl}{0.267} & 11.33 & 66.0 \\
OctreeGS & \cellcolor{orangehl}{26.43} & \cellcolor{redhl}{0.771} & \cellcolor{orangehl}{0.417} & \cellcolor{yellowhl}{5.13} & \cellcolor{yellowhl}{125.0} & \cellcolor{redhl}{23.81} & \cellcolor{redhl}{0.758} & \cellcolor{redhl}{0.295} & \cellcolor{yellowhl}{6.78} & \cellcolor{yellowhl}{83.0} & \cellcolor{orangehl}{22.31} & \cellcolor{orangehl}{0.793} & \cellcolor{orangehl}{0.279} & \cellcolor{yellowhl}{5.58} & \cellcolor{yellowhl}{91.0} \\
FLoD & 25.06 & 0.732 & 0.461 & \cellcolor{orangehl}{3.04} & \cellcolor{orangehl}{245.0} & 22.46 & 0.721 & 0.310 & \cellcolor{orangehl}{6.31} & \cellcolor{orangehl}{147.0} & 21.57 & 0.761 & 0.305 & \cellcolor{orangehl}{4.02} & \cellcolor{orangehl}{174.0} \\
FilterGS & \cellcolor{yellowhl}{26.31} & \cellcolor{yellowhl}{0.763} & \cellcolor{yellowhl}{0.433} & \cellcolor{redhl}{1.14} & \cellcolor{redhl}{372.0} & \cellcolor{yellowhl}{23.07} & \cellcolor{orangehl}{0.749} & \cellcolor{yellowhl}{0.308} & \cellcolor{redhl}{1.01} & \cellcolor{redhl}{234.0} & \cellcolor{yellowhl}{22.13} & \cellcolor{yellowhl}{0.786} & \cellcolor{yellowhl}{0.283} & \cellcolor{redhl}{0.93} & \cellcolor{redhl}{297.0} \\

\noalign{\vspace{2pt}}
\hline
\noalign{\vspace{2pt}}
\multirow{2}{*}{\textbf{Method}} & 
\multicolumn{5}{c|}{\textbf{College}~\cite{gauuscene}} & 
\multicolumn{5}{c|}{\textbf{Residence}~\cite{gauuscene}} & 
\multicolumn{5}{c}{\textbf{Modern-Building}~\cite{gauuscene}} \\

 & \rotatebox{0}{\fontsize{8}{10}\selectfont PSNR$\uparrow$} 
 & \rotatebox{0}{\fontsize{8}{10}\selectfont SSIM$\uparrow$} 
 & \rotatebox{0}{\fontsize{8}{10}\selectfont LPIPS$\downarrow$} 
 & \rotatebox{0}{\fontsize{9}{10}\selectfont $t_\text{f}$ $\downarrow$} 
 & \rotatebox{0}{\fontsize{8}{10}\selectfont FPS$\uparrow$} 
 & \rotatebox{0}{\fontsize{8}{10}\selectfont PSNR$\uparrow$} 
 & \rotatebox{0}{\fontsize{8}{10}\selectfont SSIM$\uparrow$} 
 & \rotatebox{0}{\fontsize{8}{10}\selectfont LPIPS$\downarrow$} 
 & \rotatebox{0}{\fontsize{9}{10}\selectfont $t_\text{f}$ $\downarrow$} 
 & \rotatebox{0}{\fontsize{8}{10}\selectfont FPS$\uparrow$} 
 & \rotatebox{0}{\fontsize{8}{10}\selectfont PSNR$\uparrow$} 
 & \rotatebox{0}{\fontsize{8}{10}\selectfont SSIM$\uparrow$} 
 & \rotatebox{0}{\fontsize{8}{10}\selectfont LPIPS$\downarrow$} 
 & \rotatebox{0}{\fontsize{9}{10}\selectfont $t_\text{f}$ $\downarrow$} 
 & \rotatebox{0}{\fontsize{8}{10}\selectfont FPS$\uparrow$} \\
\hline

3DGS & 24.21 & 0.697 & 0.341 & -- & 58.0 & 24.15 & 0.746 & 0.250 & -- & 42.0 & 25.79 & 0.733 & 0.324 & -- & 54.0 \\
H3DGS & 24.79 & 0.713 & 0.330 & \cellcolor{yellowhl}{5.43} & 82.0 & 24.76 & 0.751 & 0.236 & 9.14 & 61.0 & 25.93 & 0.748 & 0.391 & 7.26 & 89.0 \\
LoG & \cellcolor{redhl}{25.9} & \cellcolor{redhl}{0.757} & \cellcolor{redhl}{0.282} & 11.97 & 72.0 & \cellcolor{yellowhl}{25.09} & \cellcolor{yellowhl}{0.781} & \cellcolor{yellowhl}{0.223} & 14.19 & 57.0 & \cellcolor{redhl}{27.35} & \cellcolor{redhl}{0.814} & \cellcolor{redhl}{0.270} & 8.51 & 81.0 \\
OctreeGS & \cellcolor{yellowhl}{25.62} & \cellcolor{yellowhl}{0.746} & \cellcolor{orangehl}{0.287} & 6.04 & \cellcolor{yellowhl}{104.0} & \cellcolor{redhl}{25.37} & \cellcolor{orangehl}{0.787} & \cellcolor{redhl}{0.218} & \cellcolor{yellowhl}{7.59} & \cellcolor{yellowhl}{78.0} & \cellcolor{yellowhl}{26.56} & \cellcolor{yellowhl}{0.792} & \cellcolor{yellowhl}{0.284} & \cellcolor{yellowhl}{5.59} & \cellcolor{yellowhl}{120.0} \\
FLoD & 25.28 & 0.701 & 0.329 & \cellcolor{orangehl}{3.79} & \cellcolor{orangehl}{183.0} & 24.2 & 0.743 & 0.252 & \cellcolor{orangehl}{5.11} & \cellcolor{orangehl}{137.0} & 26.02 & 0.742 & 0.311 & \cellcolor{orangehl}{2.13} & \cellcolor{orangehl}{203.0} \\
FilterGS & \cellcolor{orangehl}{25.69} & \cellcolor{orangehl}{0.748} & \cellcolor{yellowhl}{0.288} & \cellcolor{redhl}{1.05} & \cellcolor{redhl}{290.0} & \cellcolor{orangehl}{25.31} & \cellcolor{redhl}{0.789} & \cellcolor{orangehl}{0.219} & \cellcolor{redhl}{1.81} & \cellcolor{redhl}{212.0} & \cellcolor{orangehl}{27.04} & \cellcolor{orangehl}{0.810} & \cellcolor{orangehl}{0.273} & \cellcolor{redhl}{1.04} & \cellcolor{redhl}{354.0} \\

\noalign{\hrule height 0.8pt}
\end{tabularx}
\caption{\textbf{Quantitative results on MatrixCity\cite{matrixcity}, UrbanScene\cite{urbanscene} and GauUScene\cite{gauuscene} dataset.} The \colorbox{redhl}{1st}, \colorbox{orangehl}{2nd} and \colorbox{yellowhl}{3rd} best results are highlighted.
    Our method achieves state-of-the-art rendering FPS by reducing filter time $t_f$(ms) and employing the Gaussian shrinking strategy, maintaining high-fidelity reconstruction quality competitive with the top-performing approaches.}
\label{tab:4-1}
\end{table*}

\heading{Gaussian Shrinking Strategy.} 
Our goal is to adaptively shrink Gaussians to mitigate congestion without compromising visual fidelity. For a 2D Gaussian, reducing its effective influence radius $r$ based on its central opacity $\alpha_0$ is a direct approach. Given the relationship between a Gaussian's opacity and its spatial decay, the radius $r$ at which the opacity decays to a threshold $\tau$ can be derived as $r = \sqrt{2\sigma_{\max} \ln(\alpha_0 / \tau)}$.

The challenge lies in setting $\tau$. A fixed threshold $\tau=1/255$, as used in FlashGS is overly conservative for congested scenes, leaving many redundant Gaussians untouched. To achieve a better trade-off, we propose a scene-adaptive threshold $\tau$ that leverages our GTC metric.

The core intuition is that the tile-wise GTC ($\bar{G}_v$) represents the average \textit{contribution budget} per Gaussian. A low $\bar{G}_v$ indicates a highly congested view, necessitating a more aggressive pruning strategy to restore efficiency. Unlike fixed thresholds, GTC captures spatial variations in Gaussian utility and enables content-aware adjustment. This scene-adaptive logic motivates our design of the shrinking threshold $\tau$ to be inversely proportional to the average GTC: 
\begin{equation}
    \bar{G} = \frac{1}{N} \sum_{j=1}^{N} \bar{G_j},~~\tau = \lambda_G \cdot \bar{G}^{-1}
    \label{eqa:define-tau}
\end{equation}
where $N$ is the number of images and $\lambda_G$ is a scaling factor that controls the overall aggressiveness of shrinking. This formulation is the algorithmic novelty of our strategy: it automatically increases the pruning aggressiveness (higher $\tau$) in scenes with severe Gaussian congestion (low $\bar{G}$), yielding a superior efficiency-quality trade-off compared to fixed-threshold methods.

\section{Experiment}
\subsection{Setup}
\vspace{-0.5em}
\heading{Dataset.}
We conduct experiments on large-scale datasets, including the synthetic dataset MatrixCity\cite{matrixcity}, real-world datasets GauUScene\cite{gauuscene} and UrbanScene\cite{urbanscene}. For MatrixCity, we use the drone-captured \textit{Block Small} subset containing 5,620 images, covering an area of 2.7 $km^2$. In GauUScene, we evaluate on the \textit{College}, \textit{Residence} and \textit{Modern-Building} scenes while for UrbanScene we select the \textit{Residence} and \textit{Sci-art} scenes, all covering over 1 $km^2$ each. 
Camera poses are estimated using COLMAP\cite{colmap}.

\vspace{-0.5em}
\heading{Metric.}
We adopt the PSNR, SSIM\cite{ssim}, and LPIPS\cite{lpips} to evaluate novel view synthesis quality. The term $t_\text{f}$(ms) represents the time expended on Gaussian filtering.
To assess rendering speed, we 
report the average FPS derived from the cumulative time across all frames.

\vspace{-0.5em}
\heading{Baseline.}
We compare our method with vanilla 3DGS\cite{3dgs} and the following LoD-based methods: H3DGS\cite{hieGS}, FLoD\cite{flod}, LoG\cite{LoG} and OctreeGS\cite{octreegs}. These LoD methods are specifically designed for large-scale scene reconstruction by constructing hierarchical Gaussian structures, with a primary focus on accelerating the rendering process for such extensive environments. 

\vspace{-0.5em}
\heading{Implementation.}
Our method is trained for 100–300k iterations depending on the number of input images, while all baselines are trained for the same number of iterations using their respective default settings. 
The training process is conducted on a Nvidia A100(40G), and evaluation is performed on a RTX 4090. In the Gaussian shrinking strategy, we set $\lambda_G = 0.2$. All renderings are in 1080p resolution.

\subsection{Comparison}

As shown in Table~\ref{tab:4-1}, FilterGS outperforms all baseline methods in reducing the Gaussian preparation time $t_f$ prior to rasterization. Gaussian preparation is a major bottleneck for rendering speed of LoD-based methods, so this reduction leads to a significant FPS increase, achieving the fastest rendering speed across all tested datasets. Furthermore, while aiming for the highest rendering speed, our method achieves competitive reconstruction quality among LoD-based approaches, ensuring the faithful preservation of fine details.

To compare different Gaussian shrinking strategies, we contrast our approach with existing methods: vanilla 3DGS uses the standard $3\sigma$ principle, while FlashGS employs a fixed opacity threshold ($\tau=1/255$). In contrast, our shrinking threshold $\tau$ is scene-adaptive, determined by the GTC metric, with the scaling factor $\lambda_G=0.2$. We also analyze the total number of generated key-value pairs ($N_P$). 

\begin{figure}[bp]
\begin{subfigure}[t]{0.48\textwidth}
    \centering
    \fontsize{9}{11}\selectfont
    \setlength{\tabcolsep}{3pt}
    \renewcommand{\arraystretch}{1.1}
    
    \begin{tabular}{@{}cc|cccccc@{}}
    \toprule
     & & \multicolumn{3}{c}{\textbf{College}} & \multicolumn{3}{c}{\textbf{Modern}} \\
    \multicolumn{2}{c|}{\textbf{shrinking}} &
    \fontsize{9}{10}\selectfont PSNR & 
    \fontsize{9}{10}\selectfont FPS & 
    \fontsize{8}{10}\selectfont ${N_P}$$\downarrow$ & 
    \fontsize{9}{10}\selectfont PSNR & 
    \fontsize{9}{10}\selectfont FPS & 
    \fontsize{8}{10}\selectfont $N_P$$\downarrow$ \\
    \midrule
    3DGS & 3$\sigma$ & 
    24.21 & 58 & 7.99M & 25.79 & 64 & 5.82M \\
    3DGS & $\tau$=1/255 & 
    24.21 & 61 & 7.98M & 25.79 & 66 & 5.81M \\
    3DGS & $\lambda_G$=0.2 & 
    24.07 & 72 & 6.20M & 25.67  & 80 & 4.31M \\
    FilterGS & 3$\sigma$ & 
    \textbf{25.90} & 241 & 3.25M & \textbf{27.35} & 302 & 2.13M \\
    FilterGS & $\tau$=1/255 & 
    25.90 & 246 & 3.24M & 27.35 & 307 & 2.12M \\
    FilterGS & $\lambda_G$=0.2 & 
    25.69 & \textbf{290} & \textbf{1.56M} & 27.04 & \textbf{354} & \textbf{1.12M} \\
    \bottomrule
    \end{tabular}
    \vspace{0.5em} 
\end{subfigure}
\begin{subfigure}[t]{0.48\textwidth}
  \centering
   \includegraphics[width=\linewidth]{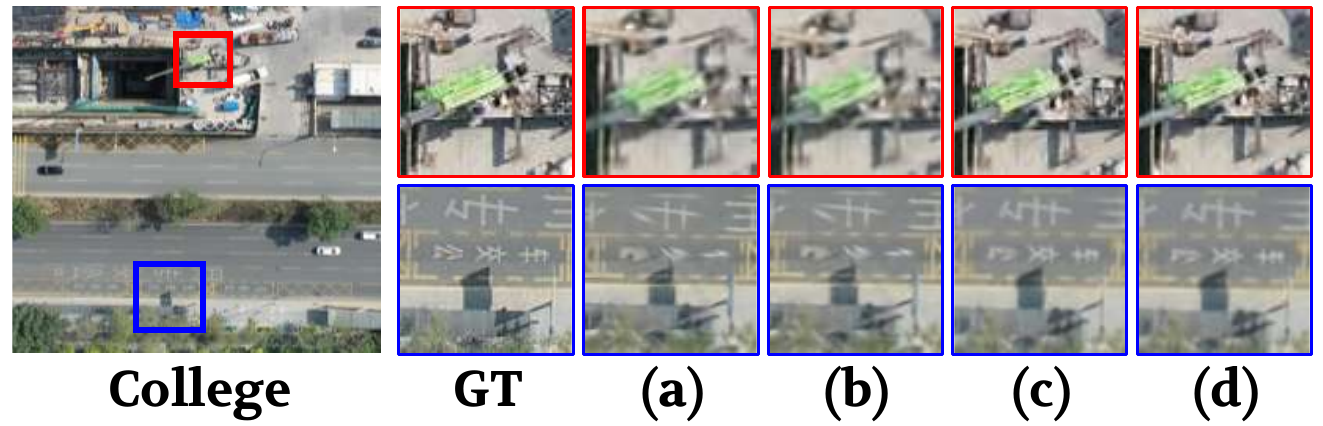}
\end{subfigure}
\caption{\textbf{Comparison of Different Shrinking Strategy on College and Modern-Building\cite{gauuscene}.} The best results are highlighted \textbf{in bold}. $N_P$ represents the total number of generated key-value pairs. (a) to (d) show the rendered detail: (a) vanilla 3DGS ($3\sigma$ principle); (b) 3DGS with our shrinking strategy ($\lambda_G=0.2$); (c) FilterGS with the $3\sigma$ principle; (d) FilterGS with $\lambda_G=0.2$. }
\label{fig:4-2}
\end{figure}

As shown in Figure~\ref{fig:4-2}, the hierarchical management of Gaussians in the LoD tree structure already achieves a significant reduction in $N_P$. 
Our Gaussian shrinking strategy yields a consistent 20\% FPS improvement with only a 1\% PSNR degradation in both the vanilla 3DGS and our FilterGS framework. Furthermore, our proposed redundancy elimination strategy successfully prunes an additional $75\%$ and $25\%$ of the remaining redundant key-value pairs for FilterGS and 3DGS, respectively.

Furthermore, we provide a brief analysis of the memory-performance trade-off. The introduction of the ancestor-path indexing leads to a manageable memory overhead of approximately 20\% compared to the original tree structure, with the total model size ranging between 1.61GB (Residence\cite{urbanscene})---6.33 GB (Residence\cite{gauuscene}) in our tested scenes. We argue that this is a worthwhile trade-off, as the modest increase in memory consumption enables a dramatic reduction in Gaussian filtering time and consequently a significant FPS improvement, which is often the critical bottleneck for real-time applications.

\begin{table}[bp]
    \centering
    \fontsize{9}{11}\selectfont
    \setlength{\tabcolsep}{3pt}
    \renewcommand{\arraystretch}{1.1}
    \begin{tabular}{@{}cc|ccccc|cc@{}}
    \toprule
    \fontsize{9}{10}\selectfont Shrink &
    \fontsize{9}{10}\selectfont Filter &
    \fontsize{9}{10}\selectfont $T_{\text{calcu.}}$ & 
    \fontsize{9}{10}\selectfont $T_{\text{synch.}}$ &
    \fontsize{9}{10}\selectfont $T_{\text{prepr.}}$ & 
    \fontsize{9}{10}\selectfont $T_{\text{sort~}}$ & 
    \fontsize{9}{10}\selectfont $T_{\text{alpha}}$ &
    \fontsize{9}{10}\selectfont $T_{\text{total}}$ &
    \fontsize{9}{10}\selectfont FPS  \\
    \midrule
     \textcolor{red}{\ding{55}} & \textcolor{red}{\ding{55}}
    & 3.59 & 7.72 & 0.20 & 1.04 & 2.58 & 15.13 & 66\\
     \textcolor{ForestGreen}{\checkmark} &  \textcolor{red}{\ding{55}}
    & 3.59 & 7.80 & 0.27 & 0.61 & 1.69 & 13.96 & 72\\
     \textcolor{red}{\ding{55}} &  \textcolor{ForestGreen}{\checkmark}
    & 0.52 & 0.41 & \textbf{0.19} & 1.01 & 2.64 & 4.77 & 210\\
     \textcolor{ForestGreen}{\checkmark} &  \textcolor{ForestGreen}{\checkmark} 
    &\textbf{ 0.52} & \textbf{0.40} & 0.26 & \textbf{0.57} & \textbf{1.62} & \textbf{3.37} & \textbf{297}\\
    \bottomrule
    \end{tabular}
    \vspace{-0.5em}
    \caption{\textbf{Ablation Study about Rendering Time Cost (ms) on Residence\cite{urbanscene}.} The best results are highlighted \textbf{in bold}. The total rendering time ($T_{\text{total}}$) is divided into four phases: Gaussian filtering ($T_\text{calcu.}+T_\text{synch.}$), preprocessing ($T_\text{prepr.}$), sorting ($T_\text{sort}$), and $\alpha$-blending ($T_\text{alpha}$).}
    \label{tab:2}
\end{table}

\begin{table}[bp]
    \centering
    \fontsize{9}{11}\selectfont
    \setlength{\tabcolsep}{3pt}
    \renewcommand{\arraystretch}{1.1}
\begin{tabular}{@{}cc|cccccc@{}}
    \toprule
     & & \multicolumn{3}{c}{\textbf{College}} & \multicolumn{3}{c}{\textbf{Modern}} \\
    \fontsize{9}{10}\selectfont Shrink &
    \fontsize{9}{10}\selectfont Filter &
    \fontsize{9}{10}\selectfont PSNR & 
    \fontsize{9}{10}\selectfont SSIM & 
    \fontsize{8}{10}\selectfont ${N_P}$$\downarrow$ & 
    \fontsize{9}{10}\selectfont PSNR & 
    \fontsize{9}{10}\selectfont SSIM & 
    \fontsize{8}{10}\selectfont $N_P$$\downarrow$ \\
    \midrule
    \textcolor{red}{\ding{55}} & \textcolor{red}{\ding{55}} &
    25.90 & 0.752 & 3.25M & 27.35 & 0.813 & 2.13M \\
    \textcolor{ForestGreen}{\checkmark} &  \textcolor{red}{\ding{55}} &
    25.69 & 0.748 & 1.56M & 27.04 & 0.810 & 1.12M \\
    \textcolor{red}{\ding{55}} &  \textcolor{ForestGreen}{\checkmark} &
    25.90 & 0.752 & 3.25M & 27.35 & 0.813 & 2.13M \\
    \textcolor{ForestGreen}{\checkmark} &  \textcolor{ForestGreen}{\checkmark} &
    25.69 & 0.748 & 1.56M & 27.04 & 0.810 & 1.12M \\
    \bottomrule
\end{tabular}
\vspace{-0.5em}
\caption{\textbf{Ablation Study about Rendering Quality on College\cite{gauuscene} and Modern-Building\cite{gauuscene}.} $N_P$ represents the total number of generated key-value pairs.}
\label{tab:3}
\end{table}

\subsection{Ablation Study}
\label{chap:4.3}
We systematically evaluate the contributions of our two core technical modules: the Gaussian shrinking strategy and parallel filtering strategy by measuring their independent contribution to rendering time cost and quality.

\vspace{-0.1em}
\heading{Parallel Filter Mechanism.}
To illustrate the acceleration process of this mechanism in rendering, we present the latency breakdown of different stages in Table~\ref{tab:2}, including filtering, preprocessing, sorting, and $\alpha$-blending. The filtering stage comprises both the computation of Gaussian attributes ($T_\text{calcu.}$) and the necessary synchronization process ($T_\text{synch.}$). Our parallel filter mechanism reduces the filtering time by over 90\% compared to sequential traversal method, as Table~\ref{tab:2} shown. This acceleration translates to FPS increases of 218\%. 

As shown in Table~\ref{tab:2}, the inter-level synchronization overhead of the original method accounts for over 40\% of the total rendering time. Our parallel filter mechanism reduces this synchronization time by 95\%, as it filters Gaussians across all levels simultaneously, eliminating the synchronization required by per-level traversal. 

Notably, the number of Gaussians entering the filter ($>1.5M$) far exceeds the single-batch processing limit of the RTX 4090 ($\approx0.26M$). Parallel filters ensures that each processing pass operates at full GPU utilization, thereby also reducing the total computation time for Gaussian attributes. The bottleneck of serial traversal lies in GPU synchronizations between levels and the underutilization of computational resources when Gaussians are few. Even with a tree depth of $L$ = 5, the LoG serial filtering time is 3.6× that of our parallel filter mechanism.

As shown in Table~\ref{tab:3}, the filter mechanism selects an identical set of Gaussians as the per-level traversal approach, as evidenced by the same number of generated pairs and identical reconstruction quality (PSNR, SSIM). This exact match demonstrates that our filter mechanism introduces no degradation whatsoever to rendering quality.

\heading{Gaussian Shrinking Strategy.}
As shown in Table~\ref{tab:2}, the shrinking strategy increases preprocessing time in the rendering pipeline but reduces the time cost of sorting and $\alpha$-blending by approximately 40\%, owing to the reduced number of generated key-value pairs. When combined with the parallel filter mechanism, it further improves the FPS by over 20\%.

\begin{figure}[tp]
   \centering
   \includegraphics[width=\linewidth]{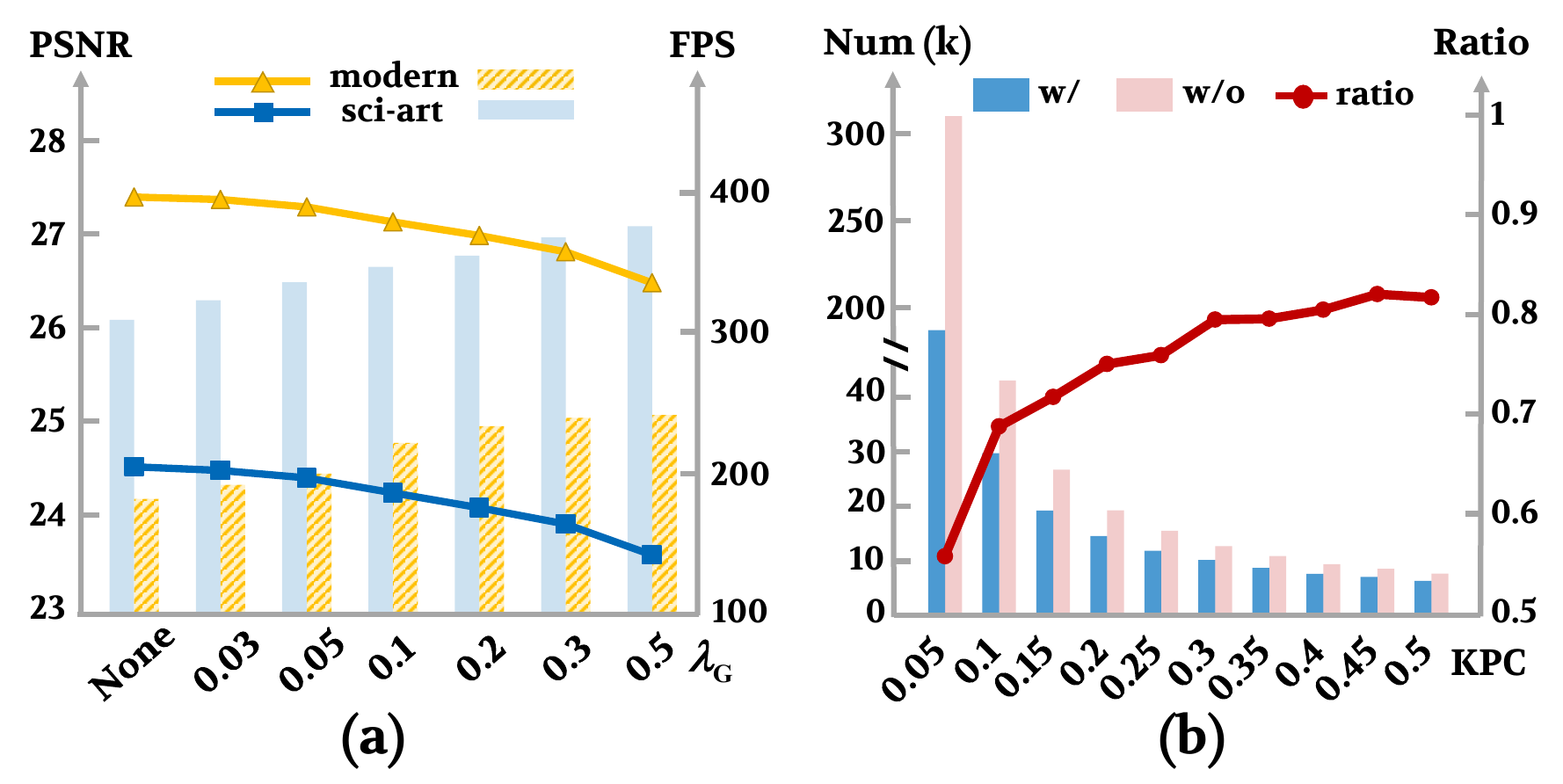}
   \vspace{-2.0em}
   \caption{\textbf{(a).} \textbf{The Trade-off between PSNR and FPS versus $\lambda_G$. ~\textbf{(b).} Distribution of Key-Value Pairs as a Function of KPC, with (w/) and without (w/o) the shrinking.} ($\lambda_G=0.2$, scene: Residence\cite{urbanscene}) }
\label{fig:4-3}
\end{figure}

The core parameter $\lambda_G$ controls the aggressiveness of the shrinking strategy. Within the $\lambda_G$ interval of [0.03, 0.2], the setting achieves a well-balanced compromise as shown in Figure~\ref{fig:4-3}(a), where a 1\% decrease in PSNR results in a 20\% frame rate improvement ($\lambda_G$=0.2). Our Gaussian shrinking approach effectively reduces KPC$<$0.05 pairs on Residence\cite{urbanscene} by over 40\% as shown in Figure~\ref{fig:4-3}(b).

\begin{figure}[tp]
\begin{subfigure}[t]{0.46\textwidth}
    \centering
    \fontsize{8}{10}\selectfont
    \setlength{\tabcolsep}{3pt}
    \renewcommand{\arraystretch}{1.1}
\begin{tabular}{@{}ccccccccc@{}}
    \toprule
    $N_{\text{low}}$ &
    \fontsize{7}{10}\selectfont $-0\%$ &
    \fontsize{7}{10}\selectfont $-10\%$ &
    \fontsize{7}{10}\selectfont $-20\%$ & 
    \fontsize{7}{10}\selectfont $-30\%$ &
    \fontsize{7}{10}\selectfont $-40\%$ & 
    \fontsize{7}{10}\selectfont $-50\%$ & 
    \fontsize{7}{10}\selectfont $-60\%$ &
    \fontsize{7}{10}\selectfont $-70\%$  \\
    \midrule
    $\lambda_G$
    & -- & 0.02 & 0.03 & 0.05 & 0.07 & 0.1 & 0.18 & 0.3\\
    $N_{\text{P}}(\text{M})$
    & 2.53 & 2.24 & 1.94 & 1.69 & 1.52 & 1.34 & 1.13 & 0.79\\
    PSNR
    & 26.52 & 26.52 & 26.51 & 26.50 & 26.48 & 26.46 & 26.43 & 26.40\\
    FPS
    & 311 & 320 & 329 & 340 & 349 & 357 & 368 & 385\\
    \midrule
    $\lambda_G$
    & -- & 0.02 & 0.03 & 0.05 & 0.09 & 0.14 & 0.29 & 0.43\\
    $N_{\text{P}}(\text{M})$
    & 3.66 & 3.25 & 2.78 & 2.46 & 2.20 & 1.94 & 1.54 & 1.33\\
    PSNR
    & 25.57 & 25.56 & 25.54 & 25.51 & 25.46 & 25.38 & 25.25 & 25.10\\
    FPS
    & 153 & 164 & 175 & 184 & 196 & 205 & 220 & 233\\
    \bottomrule
\end{tabular}
\vspace{0.1em} 
\end{subfigure}
\begin{subfigure}[t]{0.48\textwidth}
    \centering
    \includegraphics[width=\linewidth]{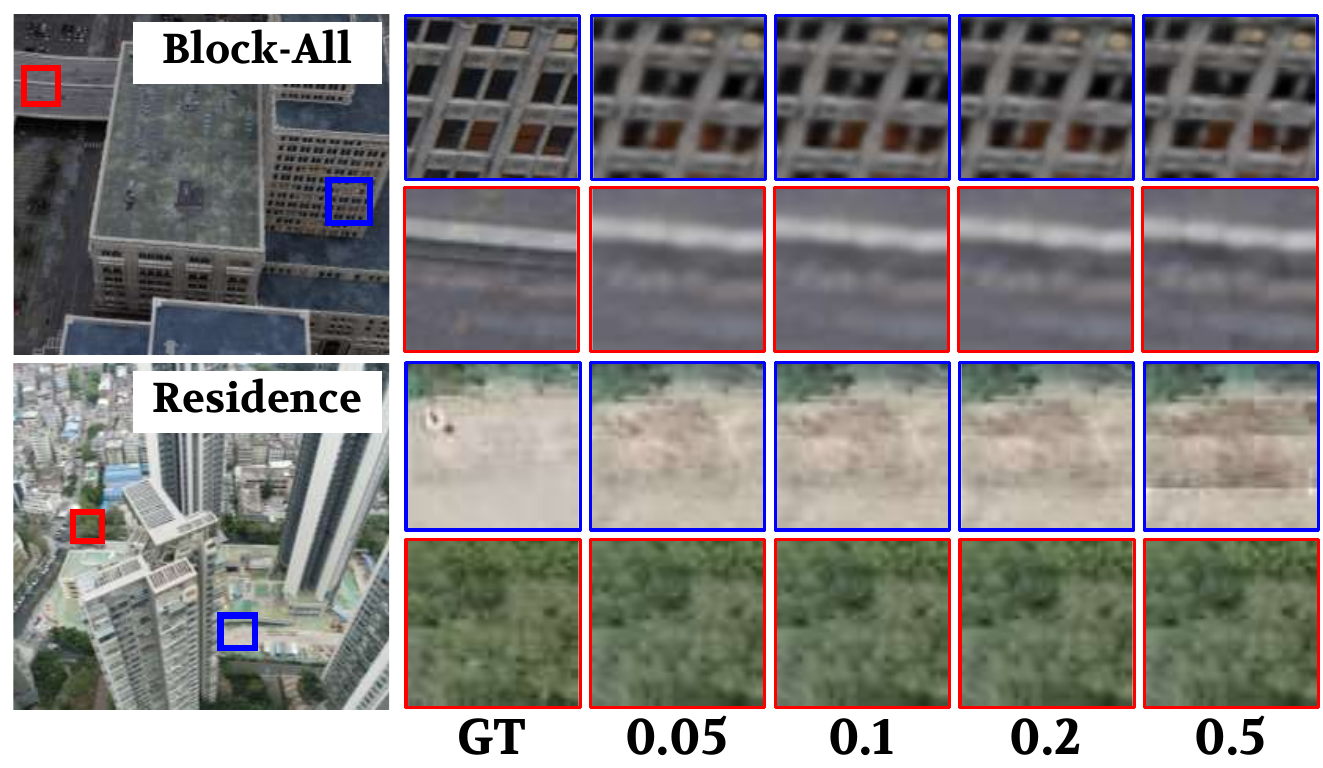}
\end{subfigure}
\vspace{-1.5em} 
\caption{\textbf{Effect of $\lambda_G$ on Key-Value Pair Pruning and Rendering Detail on Block\cite{matrixcity} and Residence\cite{gauuscene}.} $N_{\text{low}}$ represents the number (M) of KPC$<0.01$ pairs.}
\vspace{-0.8em}
\label{fig:4-4}
\end{figure}

Since our shrinking strategy preserves more than 80\% of key-value pairs with high KPC, it imposes negligible degradation on rendering quality. As observed in Figure~\ref{fig:4-4}, the shrinking strategy has minimal impact on high-frequency regions such as building facades and foliage. In contrast, for low-frequency regions (roads and sand surfaces), the tile boundaries in rendered images gradually become noticeable as $\lambda_G$ continues to increase.

High‑frequency regions are characterized by dense, structurally meaningful Gaussians with high KPC values, which are mostly preserved by our shrinking strategy. In contrast, low‑frequency regions contain numerous redundant, weakly contributing Gaussians. As $\lambda_G$ increases, aggressive shrinking removes these redundant Gaussians that originally provide smooth inter‑tile blending, leading to gradually visible tile boundaries in the rendered image.

This observation further validates the rationale of our GTC-guided shrinking: it effectively eliminates redundancy while retaining visually critical structures, and the minor degradation in smooth homogeneous regions is a reasonable trade-off for significant efficiency gains.

\section{Conclusion} 
This paper presents FilterGS, a novel method that fundamentally advances efficient rendering for large-scale LoD 3DGS models. We address the core bottlenecks of serial traversal and rendering redundancy through two key innovations. We introduce a \textit{Parallel Filtering Mechanism} to achieve fast LoD Gaussian tree filtering. Additionally, we propose a \textit{Adaptive Gaussian Shrinking Strategy} guided by our novel GTC metric, eliminating redundant Gaussian-tile pairs. Experiments demonstrate that FilterGS achieves state-of-the-art FPS while maintaining competitive reconstruction quality.

\section*{Acknowledgement}
This work was partly supported by National Natural Science Foundation of China (Grant No. NSFC 62233002 and 92370203). The authors would like to thank all other members of ININ Lab of Beijing Institute of Technology for their contribution to this work.

{
    \small
    \bibliographystyle{ieeenat_fullname}
    \bibliography{main}
}


\end{document}